\documentclass[preprint,12pt]{elsarticle}

\usepackage{amssymb}
\usepackage{amsmath}

\usepackage{subcaption}

\usepackage[T1]{fontenc}
\usepackage[T5]{fontenc}

\usepackage{amsmath, amsfonts, amssymb}
\usepackage{multirow}
\usepackage{booktabs}

\usepackage{booktabs}
\usepackage{multirow}
\usepackage{array}
\usepackage{caption}
\usepackage{makecell}
\usepackage{times}
\usepackage{xcolor}
\usepackage{colortbl}
\definecolor{bestgreen}{RGB}{232,245,233}
\definecolor{groupgray}{RGB}{238,238,238}

\usepackage{arydshln}

\usepackage[ruled, lined, linesnumbered]{algorithm2e}
\SetKwComment{Comment}{/* }{ */}
\SetKwProg{Fn}{Function}{:}{end}

\usepackage[tone]{tipa}

\journal{Computers and Electrical Engineering}

\begin{document}

\begin{frontmatter}



\title{Beyond Phones: Structured Phonemic Modeling for Vietnamese Automatic Speech Recognition}


\author{Nghia Hieu Nguyen\corref{cor1}}
\ead{nghianh.nsc@grad.uit.edu.vn}

\author{Quan Ngoc Hoang}
\ead{22521178@gm.uit.edu.vn}

\author{Long Hoang Huu Nguyen}
\ead{22520817@gm.uit.edu.vn}

\author{Kiet Van Nguyen}
\ead{kietnv@uit.edu.vn}

\author{Ngan Luu-Thuy Nguyen}
\ead{ngannlt@uit.edu.vn}

\affiliation{
    organization={University of Information Technology, Vietnam National University},
    city={Ho Chi Minh city},
    country={Viet Nam}}

\cortext[cor1]{Corresponding author}

\begin{abstract}

Phone-based representations provide a compact and acoustically grounded alternative to conventional orthographic modeling for automatic speech recognition (ASR). However, phones describe surface pronunciations and may lose lexical distinctions under dialect-dependent sound mergers, making their conversion back to orthographic text inherently ambiguous. This issue is particularly relevant to Vietnamese, where pronunciation varies considerably across regional dialects.

This work proposes a structured phonemic approach to Vietnamese ASR that moves the output representation from surface phones to abstract phonemes. Exploiting the regular phoneme-grapheme correspondence of Vietnamese, each syllable is represented by a phonemic triplet consisting of its initial, rhyme, and tone, preserving lexical distinctions while enabling deterministic reconstruction of orthographic text. We further introduce a \textbf{Phonemic Syllabic-Structure Decoder} that captures the hierarchical organization of Vietnamese syllables by first predicting the rhyme and subsequently conditioning the initial and tone predictions on the rhyme.

Experiments on the standard LSVSC and multi-dialect UIT-ViMD benchmarks demonstrate the effectiveness of the proposed approach. The best models achieve WERs of 5.83\% on LSVSC and 12.58\% on UIT-ViMD, outperforming orthographic, phonetic, and previous phonemic approaches. Further analyses reveal broader lexical coverage, reduced dependence on word-frequency patterns, and consistent behavior across Vietnamese dialects. These results demonstrate the effectiveness of moving from phonetic to structured phonemic modeling and highlight the importance of incorporating language-specific phonological structure into end-to-end ASR.

\end{abstract}




\end{frontmatter}



\section{Introduction}
\label{sec:introduction}

Automatic Speech Recognition (ASR) aims to transform acoustic speech signals into textual representations and has become a fundamental component of many human-computer interaction systems. Recent advances in end-to-end ASR have substantially simplified the conventional recognition pipeline by directly learning the mapping between speech and output sequences. Architectures based on Connectionist Temporal Classification (CTC) \cite{ctc}, attention-based encoder-decoder models \cite{sutskever2014sequence}, Transformers \cite{Vaswani2017AttentionIA}, and Conformers \cite{Conf_GulatiQCPZYHWZW20} have achieved remarkable progress across a wide range of languages. Nevertheless, the effectiveness of an ASR system depends not only on its acoustic architecture but also on the choice of output representation through which the acoustic signal is mapped to linguistic units.

Most modern ASR systems directly predict orthographic units such as characters, subwords, or words. While these representations naturally correspond to written text, the relationship between acoustic realizations and orthographic units can become complex in languages exhibiting substantial pronunciation variation. This issue is particularly relevant to Vietnamese, a tonal language with considerable regional variation across Northern, Central, and Southern dialects. The same lexical unit may exhibit different surface pronunciations across dialects, increasing the difficulty of learning a direct mapping from acoustics to orthographic forms.

Phonological representations provide an alternative by introducing an intermediate linguistic representation between speech and written text. In particular, phone-based modeling represents speech closer to its surface acoustic realization and has shown promising results for Vietnamese ASR, especially under dialectal variation. However, this advantage also introduces a fundamental limitation: \textit{phones encode surface pronunciations rather than contrastive lexical identities}. Dialect-dependent phonological processes may cause distinct lexical units to share the same surface phone sequence, making the subsequent reconstruction of orthographic text ambiguous. For example, the Vietnamese syllables ``ít'' and ``ích'', phonemically represented as /it/ and /ik/, respectively, may both be realized as [ik] in Southern Vietnamese. Once represented only by their surface phones, the lexical distinction between the two syllables is lost and cannot be uniquely recovered without additional lexical or contextual information.

This observation motivates us to move beyond surface phones and investigate \textit{phonemes} as the output representation for Vietnamese ASR. Unlike phones, which describe concrete pronunciation realizations, phonemes represent abstract contrastive categories and therefore preserve lexical distinctions despite dialect-dependent surface variation. Vietnamese is particularly suitable for such modeling because of its highly regular phoneme-grapheme correspondence. A Vietnamese syllable can be represented through three major phonological components--\textit{initial}, \textit{rhyme}, and \textit{tone}--from which its orthographic form can be systematically reconstructed. Phonemic modeling therefore provides a desirable compromise: it retains a close relationship with the acoustic structure of speech while preserving the lexical information required for orthographic transcription.

Existing phonemic approaches to Vietnamese ASR, however, typically predict the initial, rhyme, and tone using separate output heads. Although such decomposition exploits the internal composition of Vietnamese syllables, independently predicting these components overlooks their phonological dependencies. Vietnamese syllables are not arbitrary combinations of an initial, rhyme, and tone; rather, their components follow a constrained hierarchical organization in which the rhyme constitutes the phonological core and constrains the admissible combinations with other components. Consequently, treating these components as conditionally independent may produce predictions that are inconsistent with the underlying syllabic structure.

To address these limitations, we propose a \textbf{structured phonemic modeling framework for Vietnamese ASR}. First, speech transcripts are represented as sequences of phonemic syllables, with each syllable factorized into an initial, rhyme, and tone. This representation preserves lexical distinctions that may disappear at the surface-phone level while enabling deterministic reconstruction of orthographic text. Second, we introduce a \textbf{Phonemic Syllabic-Structure Decoder} that explicitly incorporates the internal organization of Vietnamese syllables. Instead of predicting the three components independently, the decoder first estimates the rhyme and subsequently conditions the initial and tone predictions on the rhyme. This dependency-aware formulation integrates linguistic structure directly into the ASR decoding process while remaining applicable to standard encoder architectures.

We evaluate the proposed framework on two complementary Vietnamese ASR benchmarks: LSVSC, representing general Vietnamese speech, and UIT-ViMD, covering speech from Northern, Central, and Southern dialects. Experiments with both Transformer and Conformer encoders show consistent improvements over orthographic, phonetic, and previous phonemic approaches. The proposed models achieve Word Error Rates (WERs) of \textbf{5.83\%} on LSVSC and \textbf{12.58\%} on UIT-ViMD. Beyond recognition accuracy, our analyses further examine phonemic-component errors, regional variation, lexical diversity, and the relationship between training-word frequency and recognition performance. The results indicate that structured phonemic modeling provides broader lexical coverage and is consistently effective across regional varieties of Vietnamese.

The main contributions of this work are summarized as follows:
\begin{itemize}
    \item We develop a phonemic representation for Vietnamese ASR in which each syllable is factorized into its initial, rhyme, and tone. The representation preserves lexical distinctions while exploiting the regular correspondence between Vietnamese phonemes and graphemes.

    \item We propose a \textbf{Phonemic Syllabic-Structure Decoder} that explicitly models dependencies among syllabic components by predicting the rhyme first and conditioning the initial and tone predictions on it, rather than treating the three components independently.

    \item We conduct extensive experiments and analyses on both standard and multi-dialect Vietnamese ASR benchmarks, demonstrating improvements in recognition accuracy and providing further evidence regarding phonemic errors, dialectal robustness, lexical diversity, and frequency bias.
\end{itemize}

\section{Related Work}

\subsection{Phonetic and Phonemic Representations for ASR}

Phonetic and phonological units have long been investigated as alternatives to orthographic output units in speech recognition. Phoneme-based systems introduce an intermediate representation between acoustic observations and lexical forms, while syllable-based modeling can reduce the mismatch between short phonetic units and longer lexical units. Previous studies have explored syllables and related subword units for large-vocabulary speech recognition~\cite{Ganapathiraju2001SyllablebasedLV}, as well as hybrid syllable-character representations for languages with strong syllabic organization~\cite{Zhang2023HybridSA}. These studies suggest that incorporating linguistically motivated units can provide useful inductive structure beyond purely orthographic tokenization.

This direction is particularly relevant to Vietnamese because regional dialects exhibit systematic pronunciation differences. Recent phonetic modeling approaches represent Vietnamese speech using surface phones rather than directly predicting orthographic tokens, allowing dialect-dependent pronunciations to be modeled closer to their acoustic realizations~\cite{hoang2026phonetic}. However, surface phones do not necessarily preserve lexical contrasts. When dialectal sound changes cause different phonemes to share the same realization, distinct lexical units may collapse into an identical phone sequence. Recovering orthographic text from such representations therefore requires resolving ambiguities that are absent from the original phonemic forms.

Phonemic modeling provides an alternative that retains phonological abstraction while preserving contrastive lexical information. ViSpeechFormer~\cite{nguyen2026vispeechformer}, for example, represents Vietnamese syllables through their initial, rhyme, and tone components and predicts these components using separate output heads. Such factorization substantially reduces the output space and exploits the internal composition of Vietnamese syllables. Nevertheless, the three components are predicted independently from a shared decoder representation, leaving their internal phonological dependencies largely unmodeled.

\subsection{End-to-End and Vietnamese Automatic Speech Recognition}

End-to-end automatic speech recognition (ASR) has progressively replaced conventional multi-stage pipelines by directly learning the mapping from acoustic observations to output sequences. Connectionist Temporal Classification (CTC) and attention-based encoder-decoder models established two major paradigms for sequence-level speech recognition, while subsequent Transformer- and Conformer-based architectures further improved acoustic modeling through self-attention and convolutional representations~\cite{chan2016las}. More recently, large-scale speech pretraining has substantially improved recognition performance, particularly in low-resource settings, by transferring acoustic knowledge learned from large speech corpora.

Research on Vietnamese ASR has followed a similar progression. Early systems primarily relied on conventional acoustic and language modeling techniques~\cite{vu2005vietnamese,le08_sltu,vu2009vietnamese}, whereas recent studies have increasingly adopted end-to-end architectures and pretrained speech models. Despite these advances, Vietnamese remains challenging because of its tonal and syllabic characteristics and substantial regional pronunciation variation. Most existing systems directly predict orthographic units such as characters, words, or subwords. Although effective for standard speech, such representations do not explicitly model the phonological structure underlying Vietnamese pronunciation.

\subsection{Syllabic Structure in Vietnamese ASR}

Vietnamese provides a particularly suitable setting for structured phonemic modeling because its syllables exhibit a highly regular internal organization. A Vietnamese syllable can be decomposed into an initial, rhyme, and tone, with the rhyme further comprising a glide, vowel nucleus, and final consonant. Rather than forming arbitrary combinations, these components interact under language-specific phonological constraints. Consequently, representing a syllable merely as a set of independently predicted components does not fully exploit its internal structure.

Existing syllable-based ASR approaches have mainly treated syllables as atomic recognition units~\cite{Ganapathiraju2001SyllablebasedLV}, combined syllable and character representations~\cite{Zhang2023HybridSA}, or decomposed syllables into multiple prediction targets~\cite{nguyen2026vispeechformer}. In contrast, the present work focuses on the \emph{dependency among phonemic components within each Vietnamese syllable}. Specifically, we model the rhyme as the phonological core and condition the prediction of the initial and tone on the rhyme. This formulation distinguishes the proposed approach from both surface-phone modeling and phonemic systems with independent prediction heads, combining lexical-preserving phonemic representations with explicit syllabic structure.

\section{Methodology} \label{sec:methodology}

Given an acoustic input sequence $X=(x_1,x_2,\ldots,x_T)$, our objective is to generate its orthographic transcription $W=(w_1,w_2,\ldots,w_N)$. Instead of directly predicting orthographic tokens or surface phones, we introduce an intermediate phonemic representation in which each Vietnamese syllable is factorized into its phonological components. The overall framework consists of two main components: a phonemic tokenization scheme that converts Vietnamese text into structured phonemic sequences and a \textbf{Phonemic Syllabic-Structure Decoder} that explicitly models dependencies among these components during recognition.

\subsection{Vietnamese Phonological Structure} \label{sec:phonological-structure}

\begin{figure}[t]
    \centering
    \includegraphics[width=0.8\linewidth]{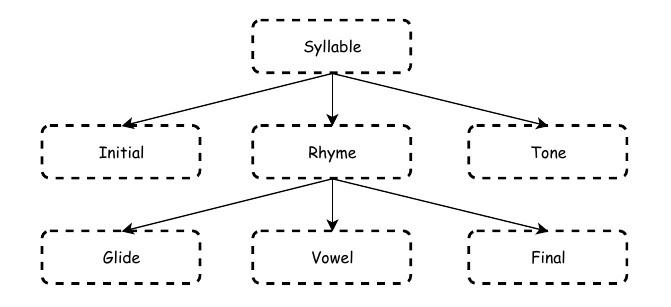}
    \caption{Phonological structure of a Vietnamese syllable.}
    \label{fig:syllabic-structure}
\end{figure}

Vietnamese syllables exhibit a highly regular phonological structure consisting of three major components: an \textit{initial}, a \textit{rhyme}, and a \textit{tone}~\cite{giap2008,giap2011} (\ref{app:phonetics-orthography}). The rhyme is compulsory and can be further decomposed into an optional glide, a mandatory vowel nucleus, and an optional final consonant~\cite{chau-dialect,haophonetic,thuat2016}, as illustrated in Figure~\ref{fig:syllabic-structure}. Thus, a syllable can be represented as
\begin{equation}
s=(i,r,t),
\end{equation}
where $i$, $r$, and $t$ denote its initial, rhyme, and tone, respectively.

Importantly, these components do not form arbitrary combinations. The rhyme constitutes the phonological core of the syllable and constrains its compatibility with the remaining components. This structured organization motivates both the proposed phonemic representation and the dependency-aware decoder introduced below.

\subsection{Phonemic Representation}
\label{sec:phonemic-representation}

Given an orthographic transcript
\begin{equation}
    W=(w_1,w_2,\ldots,w_N),
\end{equation}
we represent each syllable $w_k$ by a phonemic triplet
\begin{equation}
    p_k=(i_k,r_k,t_k),
\end{equation}
where $i_k$, $r_k$, and $t_k$ correspond to its initial, rhyme, and
tone. The complete transcript is therefore transformed into
\begin{equation}
    P=(p_1,p_2,\ldots,p_N),
\end{equation}
through a mapping
\begin{equation}
    \phi:W\rightarrow P.
\end{equation}

This representation differs fundamentally from surface-phone modeling. Phones describe concrete acoustic realizations and may therefore merge phonologically distinct units under dialectal variation. Phonemes instead encode contrastive linguistic categories and preserve the lexical identity of a syllable despite such surface variation. For example, the phonemically distinct Vietnamese syllables ``ít'' /it/ and ``ích'' /ik/ may both be realized as [ik] in Southern Vietnamese \cite{chau-dialect}. A phone representation therefore collapses this distinction, whereas their phonemic representations remain different.

Vietnamese orthography exhibits a highly regular correspondence with its phonological structure. We provided in \ref{app:phonetics-orthography} a comprehensive analysis of high correspondence between graphemes and phonemes in Vietnamese and how orthographic rules remove ambuigity while reverting phonemes back to graphemes. Consequently, the structured phonemic sequence can be mapped back to orthographic syllables through the inverse mapping
\begin{equation}
    \psi:P\rightarrow W.
\end{equation}
This property allows phonemic units to serve as the recognition targets while the final ASR output remains conventional Vietnamese text.

\subsection{Phonemic Tokenization and Reconstruction}
\label{sec:phonemic-tokenizer}

The mapping $\phi$ is implemented using a rule-based phonemic tokenization procedure. For each syllable, its phonemic sequence is retrieved and decomposed into the corresponding initial, rhyme, and tone. The rhyme is constructed by combining its glide, vowel nucleus, and final consonant. Algorithm~\ref{alg:text2phoneme} summarizes this procedure.

\SetKwFunction{FTone}{ExtractTone}
\SetKwFunction{FInitial}{ExtractInitial}
\SetKwFunction{FGlide}{ExtractGlide}
\SetKwFunction{FVowel}{ExtractVowel}
\SetKwFunction{FFinal}{ExtractFinal}
\SetKwFunction{FMerge}{MergeRhyme}

\begin{algorithm}[t]
\caption{Phonemic Tokenization of Vietnamese Text}
\label{alg:text2phoneme}
\LinesNumbered
\KwIn{A Vietnamese transcript
$W=(w_1,w_2,\ldots,w_N)$ and a phonemic dictionary $D$.}
\KwOut{A phonemic sequence
$P=(p_1,p_2,\ldots,p_N)$,
where $p_k=(i_k,r_k,t_k)$.}

$P \leftarrow [~]$\;
\ForEach{$w_k \in W$}{
    $q \leftarrow D[w_k]$\;
    $(t_k,q)\leftarrow$\FTone{$q$}\;
    $(i_k,q)\leftarrow$\FInitial{$q$}\;
    $(g_k,q)\leftarrow$\FGlide{$q$}\;
    $(v_k,q)\leftarrow$\FVowel{$q$}\;
    $c_k\leftarrow$\FFinal{$q$}\;
    $r_k\leftarrow$\FMerge{$g_k,v_k,c_k$}\;
    Append $(i_k,r_k,t_k)$ to $P$\;
}
\Return{$P$}\;
\end{algorithm}

Since the decomposition operates independently on each syllable and each helper function scans only its phonemic sequence, the procedure has linear complexity with respect to the number of input syllables, i.e., $\mathcal{O}(N)$.

At inference time, the predicted triplet $(i_k,r_k,t_k)$ is converted back into its orthographic form using $\psi$. The reconstruction is rule-based and does not employ an external language model to select among candidate words. Importantly, we impose no additional post-processing constraint on the predicted triplets. If the decoder produces an invalid initial-rhyme-tone combination, it is reconstructed directly into its corresponding literal orthographic form. Such an output consequently appears as a nonexistent or misspelled syllable and is naturally penalized by CER and WER. This setting ensures that the reported recognition performance reflects the predictions of the model rather than improvements introduced by a post-processing language constraint.

\subsection{Phonemic Syllabic-Structure Decoder}
\label{sec:phonemic-decoder}

Previous phonemic approaches decompose a Vietnamese syllable into its initial, rhyme, and tone but predict the three components independently. Such a formulation does not explicitly capture their internal phonological dependency. We therefore introduce the \textbf{Phonemic Syllabic-Structure Decoder}, which retains autoregressive decoding across syllables while explicitly modeling the structure \emph{within} each syllable.

\subsubsection{Phonemic Syllable Embedding}

At decoding step $k$, the previously generated syllable is represented by its phonemic triplet $(i_{k-1},r_{k-1},t_{k-1})$. Each component is mapped to a component-specific embedding, and the resulting vectors are concatenated and projected into the decoder hidden space:
\begin{equation}
    e_{k-1} = W_e [ E_i(i_{k-1}); E_r(r_{k-1}); E_t(t_{k-1}) ] +b_e,
\end{equation}
where $E_i$, $E_r$, and $E_t$ denote the embedding functions for the initial, rhyme, and tone, respectively. The resulting sequence is processed by the autoregressive decoder, which attends to the acoustic encoder outputs using standard multi-head attention. We denote the resulting acoustic-conditioned decoder representation at step $k$ by $f_k$.

\subsubsection{Syllabic-Structure Prediction}

A decoder with three independent output heads implicitly factorizes the prediction as
\begin{equation}
    P(i_k,r_k,t_k\mid f_k) = P(i_k\mid f_k) P(r_k\mid f_k) P(t_k\mid f_k).
\label{eq:independent}
\end{equation}
Although computationally convenient, Eq.~\eqref{eq:independent} disregards the internal organization of Vietnamese syllables.

We instead use a rhyme-first factorization:
\begin{equation}
    P(i_k,r_k,t_k\mid f_k) = P(r_k\mid f_k) P(i_k\mid f_k,r_k) P(t_k\mid f_k,r_k).
\label{eq:structured-factorization}
\end{equation}
The rhyme is selected first because it forms the phonological core of the Vietnamese syllable. Given the decoder feature $f_k$, its distribution is computed as
\begin{equation} 
P(r_k\mid f_k) = \mathrm{softmax}(W_r f_k+b_r).
    \label{eq:rhyme-prediction}
\end{equation}

The rhyme information is subsequently incorporated into the decoder representation to obtain a rhyme-aware feature:
\begin{equation}
    \tilde{f}_k = \mathcal{F}\left(f_k,E_r(r_k)\right),
    \label{eq:rhyme-aware}
\end{equation}
where $\mathcal{F}(\cdot)$ denotes the feature-fusion operation and $E_r(r_k)$ is the embedding of the predicted rhyme. The initial and tone distributions are then obtained from the shared rhyme-aware representation:
\begin{align}
    P(i_k\mid f_k,r_k) &= \mathrm{softmax}(W_i\tilde{f}_k+b_i), \\
    P(t_k\mid f_k,r_k) &= \mathrm{softmax}(W_t\tilde{f}_k+b_t).
\end{align}

This formulation assumes that, conditioned on the rhyme and acoustic-decoder representation, the initial and tone can be predicted separately. It therefore captures the principal structural dependency through the rhyme without introducing additional autoregressive positions within a syllable. The three phonemic components still constitute a single syllabic decoding step, avoiding the longer output sequence required by sequential phoneme decoding.

\subsection{Training Objective} \label{sec:training-objective}

For a target phonemic sequence $P$, the decoder is trained by maximizing the likelihood of its three syllabic components. Following the factorization in Eq.~\eqref{eq:structured-factorization}, the syllable-level negative log-likelihood is
\begin{equation}
    \mathcal{L}_{\mathrm{ph}} = -\sum_{k=1}^{N} \left[\log P(r_k\mid f_k) + \log P(i_k\mid f_k,r_k) + \log P(t_k\mid f_k,r_k) \right].
    \label{eq:phonemic-loss}
\end{equation}

When combined with a CTC branch, the complete training objective follows the standard joint CTC-attention formulation:
\begin{equation}
\mathcal{L}
=
\lambda \mathcal{L}_{\mathrm{CTC}}
+
(1-\lambda)\mathcal{L}_{\mathrm{ph}},
\label{eq:joint-loss}
\end{equation}
where $\lambda$ controls the contribution of the CTC objective.

\section{Experimental Setup} \label{sec:experimental-setup}

\subsection{Datasets} \label{sec:datasets}

We evaluate the proposed approach on two Vietnamese ASR benchmarks with complementary characteristics: LSVSC and UIT-ViMD.

\textbf{LSVSC} is a large-scale Vietnamese speech corpus designed for general-purpose speech recognition. It contains approximately 100 hours of speech from diverse speakers and recording conditions. We use the official training, development, and test partitions without introducing additional speech data. LSVSC primarily serves to evaluate the effectiveness of the proposed phonemic modeling approach under general Vietnamese speech conditions.

\textbf{UIT-ViMD} is a multi-dialect Vietnamese speech corpus covering the three major dialect regions of Vietnam: Northern, Central, and Southern. In addition to the overall test set, the corpus provides dialect-specific evaluation subsets, allowing us to investigate the robustness of different output representations under regional pronunciation variation. This dataset is particularly relevant to our study because dialect-dependent phonetic variation constitutes one of the primary motivations for moving from surface-phone to phonemic representations.

For both datasets, audio recordings are converted to mono-channel signals at 16~kHz. Non-Vietnamese words that cannot be represented by the proposed phonemic tokenizer are replaced with the \texttt{[UNK]} token. We follow the official data splits throughout all experiments.

\subsection{Compared Methods}
\label{sec:compared-methods}

We compare the proposed approach with a broad range of Vietnamese ASR systems covering different output representations and training paradigms. The compared methods can be broadly categorized into the following groups.

\textbf{Orthographic ASR.}
We consider conventional supervised Transformer and Conformer models that directly predict orthographic units. On LSVSC, we additionally compare with previously reported systems including LAS and its self-attention variants, as well as recent architectures such as ChunkFormer and other published Vietnamese ASR systems. These methods provide references for evaluating whether phonemic modeling offers advantages over direct orthographic recognition.

\textbf{Pretrained speech models.}
We include Wav2Vec2-based Vietnamese models and PhoWhisper variants, including their fine-tuned versions where available. These systems benefit from large-scale speech pretraining and therefore provide useful reference points against substantially larger pretrained models. Since the proposed models are trained from scratch using only the corresponding training sets, these comparisons should be interpreted with respect to their different training conditions.

\textbf{Phonetic ASR.}
On UIT-ViMD, we compare against previous phonetic Transformer and Conformer systems that predict surface-phone representations~\cite{hoang2026phonetic}. These systems are particularly important to our study because they explicitly model dialect-dependent acoustic realizations. Comparing against them allows us to examine whether moving from surface phones to lexical-preserving phonemes improves recognition under dialectal variation.

\textbf{Phonemic ASR.}
We further compare with ViSpeechFormer~\cite{nguyen2026vispeechformer}, which factorizes Vietnamese syllables into initial, rhyme, and tone components. Unlike our approach, these components are predicted using independent output heads. This comparison therefore helps distinguish the effect of phonemic decomposition from that of explicitly modeling dependencies among syllabic components.

Finally, we implement the proposed phonemic framework using both Transformer and Conformer backbones, denoted as \textbf{Our-Transformer} and \textbf{Our-Conformer}, respectively. Using two different encoder architectures allows us to examine whether the effectiveness of structured phonemic modeling depends on a particular acoustic backbone.

\subsection{Evaluation Metrics}
\label{sec:evaluation-metrics}

We evaluate recognition performance using \textbf{Word Error Rate (WER)} and \textbf{Character Error Rate (CER)}. Both metrics are computed on the final orthographic transcription reconstructed from the predicted phonemic sequence. WER is used as the primary metric, while CER provides a finer-grained measure of transcription errors.

To directly evaluate the intermediate phonemic representation, we additionally report \textbf{Phoneme Error Rate (PER)}. For each predicted syllable, its initial, rhyme, and tone are converted into the corresponding phoneme sequence, which is then compared with the reference phoneme sequence using the standard edit-distance formulation:

\begin{equation}
\mathrm{PER}
=
\frac{S_p+D_p+I_p}{N_p},
\end{equation}
where $S_p$, $D_p$, and $I_p$ denote the numbers of phoneme substitutions, deletions, and insertions, respectively, and $N_p$ is the total number of reference phonemes.

In addition to the overall PER, we report component-wise error rates for the \textit{initial}, \textit{rhyme}, and \textit{tone} in the subsequent analysis. These measurements allow us to identify which parts of the Vietnamese syllabic structure contribute most strongly to recognition errors.

\subsection{Implementation Details} \label{sec:implementation-details}

All proposed and supervised baseline models are implemented using the SpeechBrain toolkit~\cite{SpeechBrain} and trained from scratch on the corresponding training sets without external speech data or external language models. Speech signals are represented using 80-dimensional log-Mel filterbank features extracted with a 25-ms window and a 10-ms frame shift.

For the Transformer backbone, we use 12 encoder layers and 6 decoder layers, while the Conformer consists of 8 encoder layers and 4 decoder layers. Both architectures use a hidden dimension of 256, four attention heads, and a feed-forward dimension of 2,048. Dropout and label smoothing are both set to 0.1.

The models are optimized using Adam with the Noam learning-rate schedule. The Transformer uses a peak learning rate of $1\times10^{-3}$ with 40,000 warm-up steps, whereas the Conformer uses a peak learning rate of $4\times10^{-4}$ with 20,000 warm-up steps. Following the standard joint CTC-attention training framework~\cite{JointCTCAED_KimHW17}, the CTC loss weights are set to 0.30 and 0.15 for the Transformer and Conformer, respectively. No external language model or post-processing constraint is applied during decoding.

All experiments are conducted on an NVIDIA H100 GPU. Unless otherwise specified, the same architecture and optimization settings are used for the corresponding orthographic and phonemic models so that their differences primarily reflect the choice of output representation and decoding strategy rather than changes in model capacity or training configuration.

\section{Experimental Results} \label{sec:experimental-results}

This section evaluates the proposed structured phonemic approach on the LSVSC and UIT-ViMD benchmarks. We first compare its overall recognition performance with existing Vietnamese ASR systems and then examine the effect of explicitly modeling the internal dependency among phonemic components.

\subsection{Overall ASR Performance} \label{sec:overall-results}

\subsubsection{Results on LSVSC} \label{sec:lsvsc-results}

\begin{table}[ht]
    \centering
    \resizebox{\linewidth}{!}{
    \setlength{\tabcolsep}{5pt}
    \begin{tabular}{llrccc}
    \toprule
    \textbf{Model} & \textbf{Representation} & \textbf{\#Params} & \textbf{CER}$\downarrow$ & \textbf{WER}$\downarrow$ & \textbf{Time(s)}$\downarrow$ \\
    \midrule
    \multicolumn{6}{l}{\textit{Pretrained Orthographic Models}$^\dagger$} \\
    PhoWhisper-base (FT)    & Subword   & 244M & - & 11.23 & - \\
    PhoWhisper-medium (FT)  & Subword   & 769M & - & 10.25 & - \\
    PhoWhisper-large (FT)   & Subword   & 1.5B & - & 10.08 & - \\
    Wav2Vec2-base-vi (FT)   & Subword   & 95M  & - & 9.89  & - \\
    ZipFormer               & Subword   & 65M  & - & 10.23 & - \\
    Conv-Transformer        & Subword   & - & 7.43 & 12.59 & - \\
    ConFormer               & Subword   & - & 10.61 & 15.71 & - \\
    ZipFormer               & Subword   & - & 8.88 & 13.33 & - \\
    Multi-ConvFormer        & Subword   & - & 10.58 & 15.77 & - \\
    TASA                    & Subword   & - & 6.73 & 10.62 & - \\
    Speech Transformer      & Character & - & 6.04 & 11.16 & - \\
    \midrule
    \multicolumn{6}{l}{\textit{Fully-supervised Orthographic Models}$^\dagger$} \\
    LAS                      & Word & - & 5.79 & 9.73 & - \\
    LAS + fixedSA            & Word & - & 5.05 & 8.53 & - \\
    Transformer              & Word & 26.3M & 4.68 & 8.01 & 0.16 \\
    Transformer + fixedSA    & Word & 26.3M & 4.17 & 7.24 & 0.16 \\
    Transformer + adaptSA    & Word & 26.3M & 3.90 & 6.85 & 0.16 \\
    ChunkFormer              & Word & 110M & - & 8.85 & - \\
    \midrule
    \multicolumn{6}{l}{\textit{Phonological Models}$^\dagger$} \\
    ViSpeechFormer       & Phonemic (Independent) & - & 5.30 & 10.39 & - \\
    \midrule
    \multicolumn{6}{l}{\textit{Proposed Phonemic Models}} \\
    \textbf{Our-Transformer} & \textbf{Phonemic (Structured)} & 25.9M & \textbf{3.58} & \textbf{5.83} & 0.15 \\
    \textbf{Our-Conformer} & \textbf{Phonemic (Structured)} & 28.0M & 3.62 & 5.84 & \textbf{0.14} \\
    \bottomrule
    \end{tabular}}
    \caption{Performance comparison on the LSVSC benchmark. Lower CER and WER indicate better recognition performance. FT denotes fine-tuning of pretrained models on LSVSC. $^\dagger$Results of previously published methods are taken from \cite{gipformer,lsvsc,nguyen2026vispeechformer}.}
    \label{tab:lsvsc-results}
\end{table}

Table~\ref{tab:lsvsc-results} reports the results on LSVSC. The proposed models achieve the lowest WER and CER among the compared systems. Our-Transformer obtains a CER of \textbf{3.58\%} and a WER of \textbf{5.83\%}, while Our-Conformer achieves \textbf{3.62\%} CER and \textbf{5.84\%} WER. The comparable performance of the two architectures indicates that the improvement is not specific to a particular acoustic backbone.

Compared with conventional orthographic systems trained under a supervised setting, the proposed approach provides substantial improvements. For example, Our-Transformer reduces the WER from 8.01\% for the standard Transformer to 5.83\%, corresponding to an absolute reduction of 2.18 percentage points. It also improves upon Transformer+adaptSA, which achieves a WER of 6.85\%, by 1.02 absolute percentage points. These results demonstrate the effectiveness of incorporating phonemic structure into the recognition targets rather than directly predicting orthographic units.

The proposed models also obtain lower WERs than the reported results of several large pretrained systems, including fine-tuned PhoWhisper and Wav2Vec2 variants. However, these systems use different architectures and pretraining conditions; therefore, this comparison should be interpreted as a reference to the overall recognition performance rather than a controlled comparison of modeling strategies.

Despite the additional phonemic decomposition, the proposed approach does not introduce substantial decoding overhead. Our-Transformer and Our-Conformer require approximately 0.15 and 0.14 seconds per utterance, respectively, remaining comparable to the corresponding conventional encoder-decoder systems.

\subsubsection{Results on UIT-ViMD} \label{sec:vimd-results}

\begin{table}[ht]
    \centering
    \resizebox{\linewidth}{!}{
    \setlength{\tabcolsep}{5pt}
    \begin{tabular}{llrccccr}
    \toprule
    \textbf{Model} & \textbf{Representation} & \textbf{\#Params} & \textbf{WER}$\downarrow$ & \textbf{North} & \textbf{Central} & \textbf{South} & \textbf{Time(s)}$\downarrow$ \\
    \midrule
    \multicolumn{8}{l}{\textit{Pretrained Orthographic Models}$^\dagger$} \\
    PhoWhisper-small       & Subword & 74M  & 21.78 & 18.17 & 27.18 & 23.80 & 0.73 \\
    PhoWhisper-base        & Subword & 244M & 17.11 & 15.25 & 18.50 & 18.38 & 1.21 \\
    wav2vec2-base-vi       & Subword & 95M  & 22.48 & 19.87 & 27.15 & 23.82 & 0.03 \\
    wav2vec2-base-vi-160h  & Subword & 95M  & 31.12 & 27.09 & 40.49 & 32.83 & 0.03 \\
    wav2vec2-base-vi-250h  & Subword & 95M  & 16.83 & 14.51 & 19.73 & 18.23 & 0.07 \\
    PhoWhisper-base (FT)       & Subword & 74M & 16.30 & - & - & - & - \\
    wav2vec2-base-vi (FT)      & Subword & 95M & 15.80 & - & - & - & - \\
    wav2vec2-base-vi-160h (FT) & Subword & 95M & 17.49 & - & - & - & - \\
    wav2vec2-base-vi-250h (FT) & Subword & 95M & 13.56 & - & - & - & - \\
    \midrule
    \multicolumn{8}{l}{\textit{Fully-supervised Orthographic Models}$^\dagger$} \\
    Transformer      & Word & 29M & 16.66 & 14.68 & 18.64 & 17.95 & 0.37 \\
    Conformer        & Word & 31M & 17.56 & 16.05 & 19.92 & 18.39 & 0.29 \\
    \midrule
    \multicolumn{8}{l}{\textit{Phone-based Models}$^\dagger$} \\
    Phonetic-Transformer & Phonetic (Independent) & 29M & 13.01 & 11.26 & 15.86 & 14.14 & 0.33 \\
    Phonetic-Conformer   & Phonetic (Independent) & 31M & 13.15 & 11.29 & 16.21 & 14.13 & 0.28 \\
    \midrule
    \multicolumn{8}{l}{\textit{Proposed Phonemic Models}} \\
    \textbf{Our-Transformer} & \textbf{Phonemic (Structured)} & 26M & 12.59 & \textbf{10.77} & 14.80 & 13.68 & 0.30 \\
    \textbf{Our-Conformer} & \textbf{Phonemic (Structured)} & 28M & \textbf{12.58} & 10.88 & \textbf{14.69} & \textbf{13.58} & \textbf{0.25} \\
    \bottomrule
    \end{tabular}}
    \caption{Performance comparison on the UIT-ViMD benchmark. WER$_{\mathrm{N}}$, WER$_{\mathrm{C}}$, and WER$_{\mathrm{S}}$ denote the word error rates on the Northern, Central, and Southern dialects, respectively. FT denotes fine-tuning of pretrained models on UIT-ViMD. $^\dagger$Results of previous methods are taken from \cite{ViMD_DinhDNN24,hoang2026phonetic}.}
    \label{tab:vimd-result}
\end{table}

Table~\ref{tab:vimd-result} presents the results on the multi-dialect UIT-ViMD benchmark. Our-Transformer and Our-Conformer achieve overall WERs of \textbf{12.59\%} and \textbf{12.58\%}, respectively, providing the best performance among the compared systems.

The benefit of phonemic modeling is particularly evident when compared with conventional orthographic recognition. The supervised Transformer baseline achieves an overall WER of 16.66\%, whereas Our-Transformer reduces it to 12.59\%. Similarly, replacing the conventional output representation of the Conformer baseline reduces WER from 17.56\% to 12.58\%. Since the corresponding models use comparable backbone architectures and training conditions, these results provide direct evidence that the choice of output representation substantially affects recognition performance.

More importantly, the proposed phonemic models consistently outperform the previous phone-based approaches. Phonetic-Transformer obtains a WER of 13.01\%, compared with 12.59\% for Our-Transformer, while Phonetic-Conformer obtains 13.15\%, compared with 12.58\% for Our-Conformer. Although the absolute differences are smaller than those against orthographic baselines, the consistent improvements support our motivation for moving from surface phones to phonemes: phonetic representations capture dialectal acoustic variation but may lose lexical distinctions under pronunciation mergers, whereas phonemic representations retain these distinctions.

The improvements are also observed across all three dialect groups. Our-Transformer achieves WERs of \textbf{10.77\%}, \textbf{14.80\%}, and \textbf{13.68\%} on Northern, Central, and Southern speech, respectively, compared with 14.68\%, 18.64\%, and 17.95\% for the orthographic Transformer. Our-Conformer exhibits a similar pattern, achieving \textbf{10.88\%}, \textbf{14.69\%}, and \textbf{13.58\%}. These results indicate that the advantage of the proposed representation is consistently observed across regional varieties rather than being concentrated in a single dialect.

\subsection{Effect of Structured Phonemic Decoding} \label{sec:structured-decoder-results}

\begin{table}[ht]
    \centering
    \begin{tabular}{llcccc}
    \toprule
    \textbf{Dataset} & \textbf{Decoder} & \textbf{CER$\downarrow$} & \textbf{WER$\downarrow$} & \textbf{PER$\downarrow$} & \textbf{\#Steps} \\
    \midrule
    
    \multirow{3}{*}{LSVSC}
    & Independent & 4.12 & 6.71 & 4.44 & $N$ \\
    & Sequential  & 5.78 & 7.34 & 6.09 & $3 \times N$ \\
    & Structured  & \textbf{3.58} & \textbf{5.83} & 3.57 & $N$ \\
    
    \midrule
    
    \multirow{3}{*}{UIT-ViMD}
    & Independent & -- & 13.22 & 9.37 & $N$ \\
    & Sequential  & -- & 15.53 & 10.94 & $3 \times N$ \\
    & Structured  & -- & \textbf{12.59} & 8.21 & $N$ \\
    
    \bottomrule
    \end{tabular}
    \caption{Comparison of phonemic decoding strategies on the Transformer. All models use the same phonemic representation and training configuration.}
    \label{tab:decoder-comparison}
\end{table}

The results above establish the effectiveness of phonemic modeling, but they do not by themselves distinguish the contribution of the phonemic representation from that of the proposed syllabic-structure decoder. We therefore compare different decoding strategies under the same phonemic representation.

Specifically, we consider an \textit{independent} decoder that predicts the initial, rhyme, and tone from three separate output heads, a \textit{sequential} phoneme decoder that generates the phonemic components autoregressively, and the proposed \textit{structured} decoder that predicts the rhyme first and conditions the initial and tone on the resulting rhyme representation. The encoder architecture, phonemic targets, and training configuration are otherwise kept unchanged.

Table~\ref{tab:decoder-comparison} shows that the different phonemic decoding strategies achieve different recognition accuracy, confirming that a substantial portion of the improvement over orthographic recognition originates from the phonemic representation itself. Nevertheless, explicitly incorporating syllabic structure provides an additional advantage over treating the phonemic components independently. The structured formulation also avoids the longer autoregressive output sequence required by sequential phoneme decoding.

These findings clarify the complementary roles of the two proposed components. The phonemic representation addresses the information loss introduced by surface-phone modeling and provides the primary improvement in recognition accuracy, whereas the Phonemic Syllabic-Structure Decoder incorporates the internal organization of Vietnamese syllables without requiring each component to occupy a separate autoregressive decoding position.

\section{Analysis} \label{sec:analysis}

Beyond overall recognition accuracy, we further analyze the proposed framework from three perspectives: (1) errors across phonemic components, (2) robustness to regional pronunciation variation, and (3) lexical diversity and frequency dependence. These analyses provide further insight into how structured phonemic modeling affects Vietnamese speech recognition.

\subsection{Errors across Phonemic Components} \label{sec:component-analysis}

Since each Vietnamese syllable is represented by an initial, rhyme, and tone, we first decompose the overall Phoneme Error Rate (PER) into its three constituent components. The results are reported in Table~\ref{tab:per}.

\begin{table}[th]
    \centering
    \begin{tabular}{llrrrr}
    \toprule
    \textbf{Dataset} & \textbf{Model} & \textbf{Overall} & \textbf{Initial} & \textbf{Rhyme} & \textbf{Tone} \\
    \midrule
    \multirow{2}{*}{UIT-ViMD}
    & Transformer & 8.21 & 8.32 & 9.53 & \textbf{6.76} \\
    & Conformer   & 8.27 & 8.29 & 9.59 & 6.89 \\
    \midrule
    \multirow{2}{*}{LSVSC}
    & Transformer & 3.57 & 3.53 & 4.13 & 3.04 \\
    & Conformer   & 3.59 & 3.58 & 4.14 & \textbf{3.02} \\
    \bottomrule
    \end{tabular}
    \caption{Overall and component-wise Phoneme Error Rate (PER, \%) on UIT-ViMD and LSVSC. Lower values are better.}
    \label{tab:per}
\end{table}

A consistent pattern emerges across both datasets and acoustic backbones. Tone exhibits the lowest error rate, whereas rhyme is the most challenging component. On LSVSC, rhyme errors reach 4.13--4.14\%, compared with approximately 3\% for tone. The same pattern becomes more pronounced on UIT-ViMD, where rhyme errors increase to 9.53--9.59\%, while tone remains below 7\%.

This behavior can be partly explained by the different complexity of the three components. A rhyme may contain a glide, vowel nucleus, and final consonant and therefore represents a richer phonological unit than either the initial or tone. Moreover, the substantially larger rhyme error on UIT-ViMD suggests that this component is particularly sensitive to regional pronunciation variation. The observation is consistent across Transformer and Conformer, indicating that the error distribution is not specific to a particular acoustic backbone.

These results also provide useful context for the proposed Phonemic Syllabic-Structure Decoder. Modeling the rhyme as the central component is linguistically motivated, but it simultaneously places the most challenging phonemic component at the center of the structured prediction process. Improving rhyme modeling, particularly under regional pronunciation variation, therefore remains an important direction for further improvement.

\subsection{Robustness to Regional Variation} \label{sec:dialect-analysis}

\begin{figure}[ht]
    \centering
    \includegraphics[width=\linewidth]
    {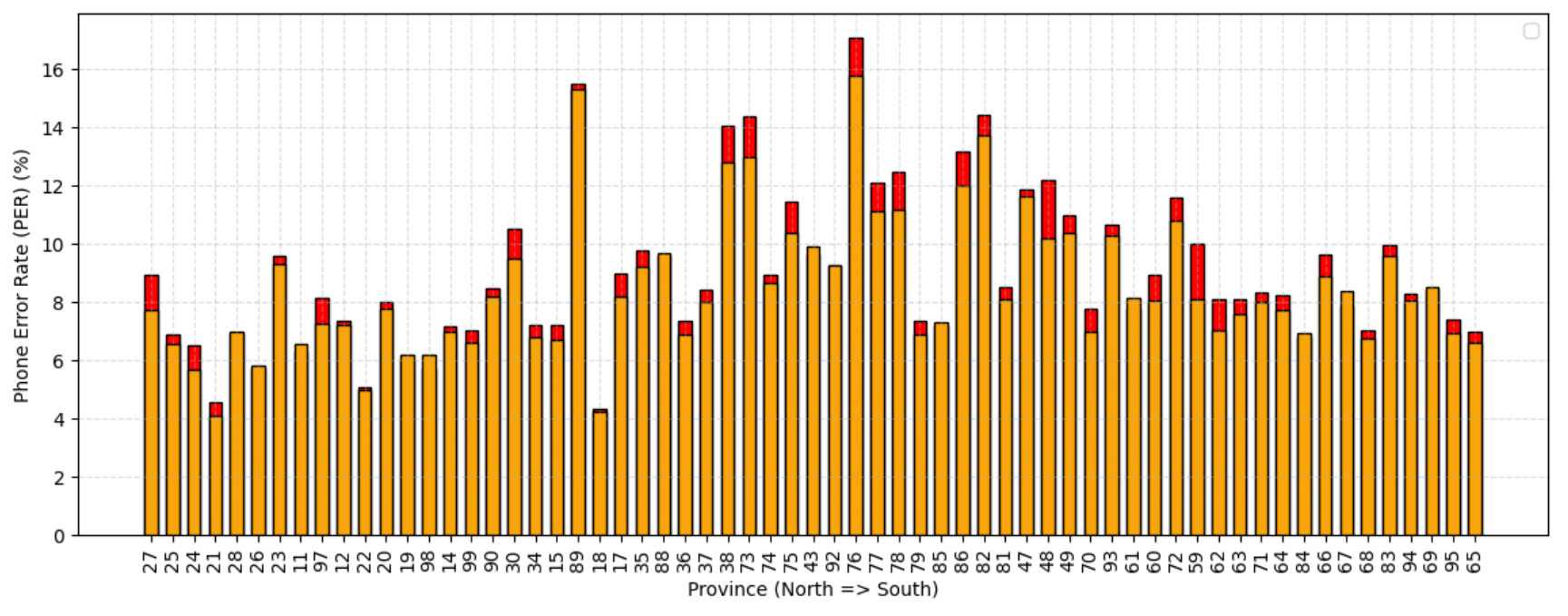}
    \caption{Phoneme Error Rate (PER) across Vietnamese provinces for the proposed Transformer and Conformer models. Provinces are ordered approximately from North to South.}
    \label{fig:per_by_prov}
\end{figure}

UIT-ViMD enables a more fine-grained analysis of recognition performance across regional varieties of Vietnamese. Figure~\ref{fig:per_by_prov} shows the PER of the proposed Transformer and Conformer models across provinces, ordered approximately from North to South.

Despite their different encoder architectures, the two models exhibit remarkably similar regional patterns. PER is relatively low for many Northern provinces, increases across the Central region, and decreases again toward the South. The close trajectories of Transformer and Conformer suggest that the observed geographical pattern is not specific to either backbone, but instead reflects differences in the difficulty of recognizing regional speech in the dataset.

This observation complements the dialect-level WER results in Table~\ref{tab:vimd-result}. Although Central speech remains the most challenging among the three dialect groups, the proposed phonemic models improve over their corresponding orthographic baselines in all regions. For Transformer, WER decreases from 14.68\% to 10.77\% in the North, 18.64\% to 14.80\% in the Central region, and 17.95\% to 13.68\% in the South. The corresponding Conformer results exhibit the same overall behavior.

Taken together, these findings indicate that the benefit of phonemic modeling is not restricted to one regional variety. At the same time, the remaining regional differences show that moving from phones to phonemes does not eliminate the difficulty introduced by dialectal pronunciation variation. Rather, phonemic modeling provides a lexically stable recognition target while the acoustic encoder must still learn the mapping from diverse surface realizations to the underlying phonological categories.

\subsection{Lexical Diversity and Frequency Dependence} \label{sec:lexical-analysis}

One motivation for phonemic modeling is that recognition is performed over reusable phonological components rather than directly over lexical forms. We therefore investigate whether this change in representation affects the diversity of correctly recognized words and the relationship between recognition performance and word frequency in the training data. Table~\ref{tab:lexical_diversity} reports the number of unique correctly recognized words together with Pearson's $r$ and Spearman's $\rho$ between training-word frequency and recognition performance.

\begin{table}[ht]
    \centering
    \small
    \begin{tabular}{lllrrr}
    \toprule
    \textbf{Dataset} &
    \textbf{Representation} &
    \textbf{Model} &
    \textbf{Unique} &
    \textbf{Pearson's $r$} &
    \textbf{Spearman's $\rho$} \\
    \midrule
    
    \multirow{4}{*}{UIT-ViMD}
    & Orthographic & Transformer & 1,696 & 0.79 & 0.76 \\
    & Orthographic & Conformer   & 1,718 & 0.77 & 0.74 \\
    & Phonemic     & Transformer & \textbf{1,983} & 0.63 & 0.54 \\
    & Phonemic     & Conformer   & 1,965 & \textbf{0.64} & \textbf{0.54} \\
    \midrule
    
    \multirow{3}{*}{LSVSC}
    & Orthographic & Transformer & 2,211 & 0.78 & 0.66 \\
    & Phonemic     & Transformer & \textbf{2,766} & \textbf{0.47} & \textbf{0.12} \\
    & Phonemic     & Conformer   & 2,748 & 0.48 & 0.14 \\
    \bottomrule
    \end{tabular}
    \caption{Lexical diversity and correlation between training-word frequency and recognition performance. Lower absolute correlation indicates weaker dependence on word frequency.}
    \label{tab:lexical_diversity}
\end{table}

Phonemic modeling consistently increases lexical diversity. On UIT-ViMD, the number of unique correctly recognized words increases from 1,696 to 1,983 for Transformer and from 1,718 to 1,965 for Conformer. On LSVSC, the phonemic Transformer recognizes 2,766 unique words correctly, compared with 2,211 for its orthographic counterpart. The increase is particularly notable because the proposed decoder does not directly predict word identities; instead, words are reconstructed from combinations of reusable phonemic components.

The correlation analysis reveals a similar pattern. Orthographic recognition exhibits a relatively strong positive association between training-word frequency and recognition performance. On LSVSC, for example, Pearson's $r$ and Spearman's $\rho$ are 0.78 and 0.66, respectively. With phonemic modeling, these values decrease to 0.47 and 0.12 for Transformer. Similar, although less pronounced, reductions are observed on UIT-ViMD.

These results indicate that the performance of the phonemic models is less strongly associated with how frequently a word occurs in the training data. Importantly, lower correlation alone does not establish that the models rely exclusively, or even predominantly, on acoustic evidence. Nevertheless, together with the increased number of correctly recognized unique words, the results are consistent with the hypothesis that factorizing lexical items into reusable phonemic components reduces dependence on word-level frequency statistics and facilitates generalization to less frequent lexical forms.

\section{Conclusion} \label{sec:conclusion}

This study investigated phonemic modeling as an alternative output representation for Vietnamese automatic speech recognition. While surface-phone representations provide a close connection to acoustic realizations, dialect-dependent sound mergers may collapse phonologically distinct lexical units into identical phone sequences. To address this limitation, we proposed a structured phonemic framework in which each Vietnamese syllable is represented by its initial, rhyme, and tone. We further introduced a \textbf{Phonemic Syllabic-Structure Decoder} that models the internal organization of Vietnamese syllables by predicting the rhyme first and conditioning the initial and tone predictions on it.

Experiments on the LSVSC and multi-dialect UIT-ViMD benchmarks demonstrate the effectiveness of the proposed approach across both Transformer and Conformer backbones. The best models achieve WERs of \textbf{5.83\%} on LSVSC and \textbf{12.58\%} on UIT-ViMD, outperforming the compared orthographic, phonetic, and previous phonemic systems. Further analyses reveal that rhyme prediction remains the primary source of phonemic errors, while the proposed representation consistently improves recognition across regional varieties. Phonemic modeling also increases the diversity of correctly recognized words and exhibits a weaker association between recognition performance and training-word frequency, suggesting improved generalization to less frequent lexical forms.

Overall, the results highlight the importance of selecting an appropriate linguistic abstraction for ASR rather than relying solely on increasingly complex acoustic architectures. For Vietnamese, phonemes provide a useful intermediate representation between surface phones and orthographic forms, combining robustness to pronunciation variation with the preservation of lexical distinctions. Future work will investigate more expressive modeling of rhyme structure, stronger mechanisms for capturing dependencies among phonemic components, and the applicability of structured phonemic modeling to other languages with systematic phoneme--grapheme correspondences.

\clearpage

\bibliographystyle{elsarticle-num-names} 
\bibliography{main}

\appendix

\section{Phonetics and Orthography in Vietnamese} \label{app:phonetics-orthography}

Traditionally, the Vietnamese had the Nom alphabet as the main orthography system. However, the Nom alphabet was developed based on the ancient Han alphabet. This means we have to be fluent in the ancient Han to have the ability to use the Nom alphabet for reading and writing. Later on, \cite{latin-viet-dictionary} used the Latin alphabet to describe the speech sound of this language. This orthography system is simple and effective for describing almost all phonetic phenomena in Vietnamese. With its advantages that counterbalance its disadvantages, this Latin alphabet is gradually improved over time and finally becomes the national alphabet system of modern Vietnamese.

Although having the Nom alphabet or the Latin alphabet, these orthography systems all reflect two consistent characteristics of Vietnamese:

\begin{enumerate}
    \item Vietnamese is a monosyllabic language.
    \item The correspondence between graphemes and phonemes in Vietnamese is consistent.
\end{enumerate}
That is, in this language, we do not have the linking pronunciation as in English, and every phoneme has persistent writing forms. In Vietnamese, each syllable has three components: initials, rhymes, and tones. Rhyme has smaller components, which are glide, vowel, and final. We provide the list of all phonemes according to the syllabic structure of Vietnamese:

\begin{itemize}
    \item 22 phonemes as the initials:
    \begin{itemize}
        \item Plosive consonants: /\textipa{b, t, t\textsuperscript{h}}, k/.
        \item Fricative consonants: /\textipa{f, d, 7, z, j, s, \:s, \t{cC}, \t{t\:s}, \ng, x, v}/.
        \item Nasal consonants: /\textipa{n, m, \ng, \textltailn}/.
        \item Vibrant consonants: /\textipa{r, l}/.
    \end{itemize}
    
    \item 01 phonemes as the glide: /\textipa{\textsubarch{u}}/.
    
    \item 15 phonemes as the vowels: 
    \begin{itemize}
        \item Diphthongs: /\textipa{ie, uo, W9}/.
        \item Monophthongs: /\textipa{a, \u{a}, \u{9}, i, E, e, u, W, o, O, O:, 9}/.
    \end{itemize}
    
    \item 10 phonemes as the finals:
    \begin{itemize}
        \item Nasal consonants: /\textipa{n, t}/.
        \item Labial consonants: /\textipa{m, p}/.
        \item Velar consonants: /\textipa{\ng, k}/.
        \item Palatal consonants: /\textipa{\textltailn, c}/.
        \item Semivowels: /\textipa{\textsubarch{u}, \textsubarch{i}}/.
    \end{itemize}

    \item 6 phonemes denote tones:
    \begin{itemize}
        \item Flat tone is denoted by nothing.
        \item Low falling tone: /\textipa{\tone{22}\tone{11}}/.
        \item Mid raising tone: /\textipa{\tone{33}\tone{55}}/.
        \item Mid falling tone: /\textipa{\tone{33}\tone{11}}/.
        \item Mid glottalized-falling tone: /\textipa{\tone{33}P\tone{11}}/.
        \item Mid glottalized-raising tone: /\textipa{\tone{33}P\tone{55}}/.
    \end{itemize}
\end{itemize}

The writing form of these phonemes is consistent regardless of grammar. In particular:

\begin{itemize}
    \item 6 tones are denoted by a mark above or below the graphemes of vowels:
    \begin{itemize}
        \item Flat tone is denoted by nothing (a).
        \item Low falling tone /\textipa{\tone{22}\tone{11}}/ is denoted by a grave accent (à).
        \item Mid raising tone /\textipa{\tone{33}\tone{55}}/ is denoted by an acute accent (á).
        \item Mid falling tone /\textipa{\tone{33}\tone{11}}/ is denoted by a hook above (ả).
        \item Mid glottalized-falling tone /\textipa{\tone{33}P\tone{55}}/ is denoted by a tilde above (ã).
        \item Mid glottalized-raising tone /\textipa{\tone{33}P\tone{11}}/ is denoted by a dot below (ạ).
    \end{itemize}
    In the following texts, we provide writing forms of vowels regarding the mentioned phonemes. These examples might include the tone mark above or below the graphemes. Readers can discard these marks to see the true writing form of the vowels in Vietnamese. For instance, the dot below \textbf{iê} /\textipa{i9}/ in word \textbf{kiệm} /\textipa{kiem\tone{33}P\tone{55}}/ denotes the mid glottalized-raising tone /\textipa{\tone{33}P\tone{55}}/.
    
    \item 22 initials have 26 writing forms:
    \begin{itemize}
        \item b /\textipa{b}/. Eg: \textbf{b}a mẹ, \textbf{b}ánh kẹo, \textbf{b}uôn \textbf{b}án.
        \item t /\textipa{t}/. Eg: \textbf{t}âm \textbf{t}ư, \textbf{t}ịnh \textbf{t}iến, \textbf{t}ính cách.
        \item th /\textipa{t\textsuperscript{h}}/. Eg: \textbf{th}ách \textbf{th}ức, \textbf{th}ành \textbf{th}ạo.
        \item k, c, or q /\textipa{k}/. Eg: \textbf{c}ách mạng, \textbf{q}uan hệ, hiện \textbf{k}im.
        \item ph /\textipa{f}/. Eg: \textbf{ph}ụ huynh, \textbf{ph}ong cách, \textbf{ph}ân định.
        \item đ /\textipa{d}/. Eg: \textbf{đ}ưa \textbf{đ}ón, \textbf{đ}ậm \textbf{đ}à.
        \item gh or g /\textipa{7}/. Eg: \textbf{g}a tàu, \textbf{g}ánh hát, \textbf{g}anh \textbf{gh}ét.
        \item gi /\textipa{z}/. Eg: \textbf{gi}ếng nước, \textbf{gi}ống loài.
        \item đ /\textipa{d}/. Eg: \textbf{đ}ứng \textbf{đ}ắn, \textbf{đ}êm tối, \textbf{đ}èn \textbf{đ}uốc.
        \item d /\textipa{j}/. Eg: \textbf{d}a \textbf{d}ẻ, tiêu \textbf{d}ùng, \textbf{d}ụng cụ.
        \item x /\textipa{s}/. Eg: sản \textbf{x}uất, \textbf{x}uất \textbf{x}ứ, \textbf{x}e cộ.
        \item s /\textipa{\:s}/. Eg: xác \textbf{s}uất, \textbf{s}o \textbf{s}ánh, \textbf{s}ao chép.
        \item ch /\textipa{\t{cC}}/. Eg: \textbf{ch}ứa \textbf{ch}an, \textbf{ch}e \textbf{ch}ở/.
        \item tr /\textipa{\t{t\:s}}/. Eg: \textbf{tr}anh chấp, tiệt \textbf{tr}ùng.
        \item ng or ngh /\textipa{ng}/. Eg: mong \textbf{ng}óng, tình \textbf{ngh}ĩa.
        \item nh /\textipa{\textltailn}/. Eg: \textbf{nh}à cửa, nỗi \textbf{nh}ớ, \textbf{nh}ung lụa.
        \item l /\textipa{l}/. Eg: \textbf{l}ấm \textbf{l}em, \textbf{l}ung \textbf{l}inh, \textbf{l}ối về.
        \item r /\textipa{r}/. Eg: \textbf{r}ậm \textbf{r}ạp, \textbf{r}ón \textbf{r}én, \textbf{r}ực \textbf{r}ỡ.
        \item kh /\textipa{x}/. Eg: \textbf{kh}ó \textbf{kh}ăn, \textbf{kh}ởi sắc, \textbf{kh}ấm \textbf{kh}á.
        \item v /\textipa{v}/. Eg: \textbf{v}ui \textbf{v}ẻ, \textbf{v}ương \textbf{v}ấn, \textbf{v}ẫy \textbf{v}ùng.
        \item m /\textipa{m}/. Eg: \textbf{m}ong \textbf{m}ỏi, \textbf{m}ay \textbf{m}ắn, \textbf{m}ênh \textbf{m}ông.
        \item n /\textipa{n}/. Eg: đất \textbf{n}ước, \textbf{n}úi \textbf{n}on, \textbf{n}ông cạn.
    \end{itemize}

    \item The glide /\textipa{\textsubarch{u}}/ has two writing forms: u or o. Eg: q\textbf{u}ê nhà, h\textbf{o}a cỏ, kh\textbf{u}yến khích.

    \item 03 diphthongs have 8 writing forms:
    \begin{itemize}
        \item iê, yê, ia, or ya /\textipa{ie}/. Eg: kh\textbf{iế}m thị, \textbf{yê}n ắng, ch\textbf{ia} sẻ, khu\textbf{ya} khoắt.
        \item uô or ua /\textipa{uo}/. Eg: kh\textbf{uô}n khổ, m\textbf{ua} bán.
        \item ươ or ưa /\textipa{W@}/. Eg: kh\textbf{ướ}u giác, dây d\textbf{ưa}.
    \end{itemize}

    \item 12 monophthongs have 13 writing forms:
    \begin{itemize}
        \item a /\textipa{a}/. Eg: b\textbf{a} mẹ, tr\textbf{a}nh vẽ, ng\textbf{ã} b\textbf{a}, m\textbf{ải} miết, làng ch\textbf{ài}.
        \item ă or a /\textipa{ă}/. Eg: ánh \textbf{ắ}ng, n\textbf{ắ}m tay, n\textbf{ă}m tháng, \textbf{áy} n\textbf{áy}, chạy nhảy.
        \item â /\textipa{\u{@}}/. Eg: n\textbf{â}ng niu, \textbf{ấ}n định, ng\textbf{â}n vang.
        \item i or y /\textipa{i}/. Eg: th\textbf{i} cử, tr\textbf{ĩ}u nặng, b\textbf{ĩ}u môi.
        \item ê /\textbf{e}/. Eg: k\textbf{ế}t quả, thể hi\textbf{ệ}n, mân m\textbf{ê}.
        \item e /\textipa{E}/. Eg: mùa h\textbf{è}, x\textbf{e} cộ, t\textbf{é} ngã.
        \item u /\textipa{u}/. Eg: th\textbf{u} mua, m\textbf{ủ}m mỉm, l\textbf{u}ng lay, tr\textbf{u}ng thành.
        \item ư /\textipa{W}/. Eg: tr\textbf{ư}ng cầu, xây d\textbf{ự}ng, \textbf{ư}ng ý.
        \item o /\textipa{O}/. Eg: chăm s\textbf{ó}c, m\textbf{o}ng ng\textbf{ó}ng, tr\textbf{o}ng veo.
        \item oo /\textipa{O:}/. Eg: x\textbf{oo}ng chảo.
        \item ô /\textipa{o}/. Eg: tr\textbf{ố}ng đ\textbf{ồ}ng, \textbf{ố}ng hút, c\textbf{ố} nhân.
        \item ơ /\textipa{@}/. Eg: m\textbf{ơ} mộng, c\textbf{ơ} nh\textbf{ỡ}, ch\textbf{ơ}i b\textbf{ờ}i.
    \end{itemize}

    \item 10 final consonants have 12 writing forms:
    \begin{itemize}
        \item i or y /\textipa{\textsubarch{i}}/. Eg: làng chà\textbf{i}, mỏ\textbf{i} mệt, chạ\textbf{y} đua, bay nhả\textbf{y}.
        \item m /\textipa{m}/. Eg: ê\textbf{m} ấ\textbf{m}, nhiệ\textbf{m} màu, mâ\textbf{m} cỗ.
        \item n /\textipa{n}/. Eg: na\textbf{n} giải, no\textbf{n} nớt, tả\textbf{n} mạ\textbf{n}.
        \item ng /\textipa{\ng}/. Eg: sa\textbf{ng} trọ\textbf{ng}, trố\textbf{ng} trải, su\textbf{ng} túc.
        \item nh /\textipa{\textltailn}/. Eg: nha\textbf{nh} nhẹn, bê\textbf{nh} vực, bi\textbf{nh} quyền.
        \item p /\textipa{p}/. Eg: phậ\textbf{p} phồng, thấ\textbf{p} thỏm, thá\textbf{p} tùng, gượng é\textbf{p}, ức hiế\textbf{p}, tẩm ướ\textbf{p}.
        \item t /\textipa{t}/. Eg: lấn á\textbf{t}, bá\textbf{t} đĩa, kế\textbf{t} quả, hớ\textbf{t} hả.
        \item c /\textbf{k}/. Eg: cú\textbf{c} áo, chự\textbf{c} chờ, bố\textbf{c} vá\textbf{c}.
        \item ch /\textbf{c}/. Eg: cá\textbf{ch} thứ\textbf{c}, chí\textbf{ch} ngừa.
        \item u or o /\textipa{\textsubarch{u}}/. Eg: tra\textbf{u} chuốt, chị\textbf{u} đựng, xoong chả\textbf{o}, xiêu vẹ\textbf{o}.
    \end{itemize}
    
\end{itemize}

\begin{figure}
    \centering
    \includegraphics[width=0.75\linewidth]{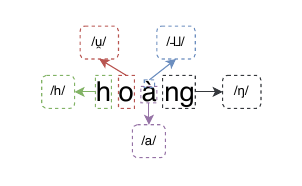}
    \caption{Example for the consistency between graphemes and phonemes in Vietnamese.}
    \label{fig:syllable-example}
\end{figure}

It is important to note that the writing form of phonemes in Vietnamese is consistent in all usage cases, regardless of grammar (tense, aspect, mood). This feature defines Vietnamese as an isolating language whose morphology is an aspect of phonetics rather than grammar, as in inflectional languages. To this end, given a Vietnamese word, native speakers can find no difficulty in pronouncing it. On the other hand, Vietnamese natives can write down any Vietnamese word by listening to its speech sound without knowing how to spell it. For instance, given the word \textbf{hoàng}, its phonetic representation is /\textipa{h\textsubarch{u}a\ng\tone{22}\tone{11}}/. We can determine the grapheme of the initial /\textipa{h}/ is \textbf{h}, of the glide /\textipa{\textsubarch{u}}/ is \textbf{o}, of the vowel /\textipa{a}/ is \textbf{a}, of the /\textipa{\ng}/ is \textbf{ng}, and tone is denoted by the grave accent above \textbf{à} (Figure \ref{fig:syllable-example}).

However, although the conversion from grapheme to phonemes is straightforward (that is, many graphemes correspond to unique phonemes), the inversion is not all ways many-to-one mapping. There are some phonemes having one-to-many mapping with graphemes, such as the initial /\textipa{7}/ can be written as \textbf{g} or \textbf{gh}, or the diphthong /\textipa{ie}/ can be written as \textbf{iê, yê, ia} or \textbf{ya}. Actually, a particular writing form of such phonemes is determined consistently via the neighbor phonemes. We detail here rules for determining writing forms of phonemes having multiple respective graphemes:

\begin{itemize}
    \item The initial /\textipa{k}/ is written as:
    \begin{itemize}
        \item \textbf{k} if followed by \textbf{i} /\textipa{i}/, \textbf{ê} /\textipa{e}/, \textbf{e} /\textipa{E}/, \textbf{iê} /\textipa{ie}/. Eg: \textbf{kị}p thời, \textbf{kiể}u cách, \textbf{kè}m cặp.
        \item \textbf{q} if followed by \textbf{u} /\textipa{\textsubarch{u}}/ as the glide. Eg: \textbf{qu}ê hương, \textbf{qu}à cáp.
        \item \textbf{c} otherwise. Eg: \textbf{cứ}ng \textbf{cỏ}i, \textbf{cở}i mở, hạt \textbf{cá}t, rau \textbf{củ}.
    \end{itemize}

    \item The initial /\textipa{7}/ is written as:
    \begin{itemize}
        \item \textbf{gh} if followed by \textbf{i} /\textipa{i}/, \textbf{ê} /\textipa{e}/, \textbf{e} /\textipa{E}/, \textbf{iê} /\textipa{ie}/. Eg: bàn \textbf{ghế}, \textbf{ghi} chép, \textbf{ghe} tàu.
        \item \textbf{g} otherwise. Eg: thanh gươm, gồng gánh, gọi điện.
    \end{itemize}

    \item The initial /\textipa{\ng}/ is written as:
    \begin{itemize}
        \item \textbf{ngh} if followed by textbf{i} /\textipa{i}/, \textbf{ê} /\textipa{e}/, \textbf{e} /\textipa{E}/, \textbf{iê} /\textipa{ie}/. Eg: \textbf{nghe} ngóng, \textbf{nghiê}m nghị, \textbf{nghệ} sĩ.
        \item \textbf{ng} otherwise. Eg: \textbf{ngà}nh nghề, \textbf{ngỗ} nghịch, \textbf{ngọ}t \textbf{ngà}o.
    \end{itemize}

    \item The diphthong /\textipa{ie}/ is written as:
    \begin{itemize}
        \item \textbf{iê} if the rhyme has a final consonant and no glide. Eg: k\textbf{iến} thức, tiết k\textbf{iệm}.
        \item \textbf{yê} if the rhyme has a final consonant and the glide written as \textbf{u}. Eg: kh\textbf{uyên} bảo, \textbf{uyển} ch\textbf{uyển}, câu ch\textbf{uyện}.
        \item \textbf{ya} if the rhyme has no final consonant and the glide written as \textbf{u}. Eg: đêm kh\textbf{uya}.
        \item \textbf{ia} if the rhyme has no final consonant and no glide. Eg: b\textbf{ìa} sách, ch\textbf{ia} sẻ.
    \end{itemize}

    \item The diphthong /\textipa{uo}/ is written as:
    \begin{itemize}
        \item \textbf{uô} if the rhyme has a final consonant. Eg: nỗi b\textbf{uồn}, m\textbf{uối} biển, m\textbf{uộn} màn, ch\textbf{uồn} ch\textbf{uồn}.
        \item \textbf{ua} if the rhyme has no final consonant. Eg: ch\textbf{ùa} chiền, nhảy m\textbf{úa}.
    \end{itemize}

    \item The diphthong /\textipa{W@}/ is written as:
    \begin{itemize}
        \item \textbf{ươ} if the rhyme has a final consonant. Eg: bia r\textbf{ượu}, h\textbf{ưởng} thụ, chiêm ng\textbf{ưỡng}.
        \item \textbf{ua} if the rhyme has no final consonant. Eg: ch\textbf{ùa} chiền, nhảy m\textbf{úa}.
    \end{itemize}

    \item The monophthong /\textipa{ă}/ is written as:
    \begin{itemize}
        \item \textbf{a} if is is followed by character \textbf{y}. Eg: m\textbf{áy} bay, c\textbf{ay} nồng, t\textbf{ay} chân.
        \item \textbf{ă} otherwise. Eg: b\textbf{ắt} tay, b\textbf{ằng} lòng, may m\textbf{ắn}.
    \end{itemize}

    \item The monophthong /\textipa{i}/ is written as:
    \begin{itemize}
        \item \textbf{i} if the rhyme has a final consonant. Eg: l\textbf{íu} r\textbf{ít}.
        \item \textbf{y} if the rhyme has no consonant. Eg: k\textbf{ỷ} luật, l\textbf{ý} do.
    \end{itemize}

    \item The final /\textipa{\textsubarch{i}}/ is written as:
    \begin{itemize}
        \item \textbf{i} if the rhyme has the front vowel /\textipa{a}/, the central vowel /\textipa{W, W9}/ or the back vowels /\textipa{u, o, O, uo}/ as the vowel. Eg: m\textbf{ải} mê, m\textbf{ui} thuyền, h\textbf{ỏi} han, m\textbf{ồi} chài, g\textbf{ửi} gắm, l\textbf{ười} biếng, n\textbf{uôi} nấng.
        \item \textbf{y} if the rhyme has /\textipa{\u{a}, \u{9}}/ as the vowel. Eg: b\textbf{ay} lượn, ch\textbf{ạy} nh\textbf{ảy}, c\textbf{ấy} c\textbf{ày}.
    \end{itemize}
\end{itemize}

Following these orthographic rules in Vietnamese, there is no ambiguity in converting phonemes to graphemes and vice versa.

\section{Appendix: Detailed Implementation of Sub-Procedures in Phonemic Tokenization Algorithm} \label{app:tokenization}

We provide here the detailed implementation of $get\_tone$, $get\_initial$, $get\_glide$, $get\_vowel$, and $get\_final$ mentioned in Section \ref{sec:phonemic-tokenizer}, Algorithm \ref{alg:text2phoneme}. According to Algorithm \ref{alg:tone}, \ref{alg:initial}, \ref{alg:glide}, \ref{alg:vowel}, and \ref{alg:final}, they defined a stable set of phonemes, then iteratively determine whether the given input $S$ is starting with any phoneme in the defined set. We can easily see that these algorithms have the linear time complexity $\mathcal{O}(n)$. Together with the algorithm \ref{alg:text2phoneme}, the overall time complexity of our tokenizer is $\mathcal{O}(n)$.

\begin{algorithm}[t]
\caption{Algorithm for determining initial of a syllable.}\label{alg:initial}
\LinesNumbered

\Fn{get\_initial($S$)}{
        \KwIn{A syllable $S$ in Vietnamese}
        \KwOut{The initial $i$ of $S$}

    $initials \leftarrow$ [\textipa{b, t, t\textsuperscript{h}, k, f, d, 7, z, j, s, \:s, \t{cC}, \t{t\:s}, \ng, x, v, n, m, \ng, \textltailn, r, l}];

    $i \leftarrow$ None;

    \For{$i$ in $initials$}
    {
        \If{$S$ starts with $i$}
        {
            Remove $i$ from the prefix of $S$;
            
            \Return{$i, S$};
        }
    }

    \Return{None, $S$}
}
\end{algorithm}

\begin{algorithm}[t]
\caption{Algorithm for determining vowel of a syllable.}\label{alg:vowel}
\LinesNumbered

\Fn{get\_vowel($S$)}{
        \KwIn{A syllable $S$ in Vietnamese}
        \KwOut{The vowel $v$ of $S$}

    $vowels \leftarrow$ [\textipa{ie, uo, W9, a, \u{a}, \u{9}, i, E, e, u, W, o, O, O:, 9, n, t, m, p, \ng, k, \textltailn, c, \textsubarch{u}, \textsubarch{i}}];

    \For{$v$ in $vowels$}
    {
        \If{$S$ starts with $v$}
        {
            Remove $v$ from the prefix of $S$;
            
            \Return{$v$, $S$};
        }
    }

    \Return{None, $S$};
}
\end{algorithm}

\begin{algorithm}[t]
\caption{Algorithm for determining final of a syllable.}\label{alg:final}
\LinesNumbered

\Fn{get\_final($S$)}{
        \KwIn{A syllable $S$ in Vietnamese}
        \KwOut{The final $f$ of $S$}

    $finals \leftarrow$ [\textipa{n, t, m, p, \ng, k, \textltailn, c, \textsubarch{u}, \textsubarch{i}}];

    \If{$f$ in $finals$}
    {
        \Return{$f$};
    }

    \Return{None};
}
\end{algorithm}

\begin{algorithm}[t]
\caption{Algorithm for determining tone of a syllable.}\label{alg:tone}
\LinesNumbered

\Fn{get\_tone($S$)}{
        \KwIn{A syllable $S$ in Vietnamese}
        \KwOut{The tone $t$ of $S$}
}
    tones $\leftarrow$ [\textipa{\tone{22}\tone{11}, \tone{33}\tone{55}, \tone{33}\tone{11}, \textipa{\tone{33}P\tone{11}}, \textipa{\tone{33}P\tone{55}}}];

    \For{$t$ in $tones$}
    {
        \If{$S$ ends with $t$}
        {
            Remove $t$ from suffix of $S$;
            
            \Return{$t$, $S$};
        }
    }

    \Return{None, $S$};
\end{algorithm}

\begin{algorithm}[t]
\caption{Algorithm for determining glide of a syllable.}\label{alg:glide}
\LinesNumbered

\Fn{get\_glide($S$)}{
        \KwIn{A syllable $S$ in Vietnamese}
        \KwOut{The glide $g$ of $S$}

    glides $\leftarrow$ [\textsubarch{u}, \textsubarch{o}];
    
    \For{$g$ in [\textipa{\textsubarch{u}, \textsubarch{o}}]}
    {
        \If{$S$ starts with $g$}
        {
            Remove $g$ from the prefix of $S$;

            \Return{\textipa{\textsubarch{u}}, $S$};
        }
    }

    \Return{None, S};
}
\end{algorithm}

\end{document}